\documentclass[11pt]{article}
\usepackage{spconf}
\usepackage{times}

\usepackage{amsmath,amssymb}
\usepackage{longtable}
\usepackage{chngpage}
\usepackage{array}
\usepackage{amssymb}
\usepackage{float}
\usepackage{booktabs}
\usepackage[pdftex]{graphicx}
\usepackage{algorithmic}
\usepackage{color}
\usepackage{graphicx}
\usepackage{subfigure}
\renewcommand{\thefigure}{\arabic{figure}}
\usepackage[algoruled,vlined]{algorithm2e}
\usepackage{longtable}

\usepackage{multirow}

\begin{document}
\title{An Adaptive Dictionary Learning Approach for \\
     Modeling Dynamical Textures}
%
\name{Xian Wei,  Hao Shen,  Martin Kleinsteuber \thanks{This work has been supported by the Cluster of Excellence CoTeSys - Cognition for Technical Systems, funded by the German Research Foundation (DFG).}}
\address{Department of Electrical Engineering and Information Technology \\
Technische Universit{\"a}t M{\"u}nchen,
Arcisstr. 21, 80333 Munich, Germany\\
\{xian.wei, hao.shen,  kleinsteuber\}@tum.de\\}
\pagestyle{empty}

\maketitle
\thispagestyle{empty}

\begin{abstract}
    Video representation is an important and challenging task in the computer vision community.
    In this paper, we assume that image frames of a moving scene can be modeled as a Linear Dynamical System. We propose a sparse coding framework, named adaptive video dictionary
    learning (AVDL), to model a video adaptively. The developed framework is
    able to capture the dynamics of a moving scene by exploring both sparse properties
    and the temporal correlations of consecutive video frames.
    The proposed method is compared with state of the art video processing methods on
    several benchmark data sequences, which exhibit appearance changes and
    heavy occlusions.
\end{abstract}

\section{Introduction}
\label{sec:01}
Temporal or dynamic textures (DT) are image sequences that exhibit spatially repetitive and certain stationarity properties in time. This kind of sequences are typically videos of processes, such as moving water, smoke, swaying trees, moving clouds, or a flag blowing in the wind. 
Study and analysis of DT is important in several applications such as video segmentation \cite{chan2009layered},
video recognition \cite{saisan2001dynamic}, and DT synthesizing \cite{doretto2003dynamic}.

One classical approach is to model dynamic scenes via the optical flow \cite{horn1981determining}. However, such methods require a certain degree of motion smoothness and parametric motion models \cite{chan2009layered}. Non-smoothness, discontinuities, and noise inherence to rapidly varying, non-stationary DTs (e.g. fire) pose a challenge to develop optical flow based algorithms.
Another technique, called particle filter \cite{djuric2003particle}, models the dynamical course of DTs as a Markov process.
A reasonable assumption in DT modeling is that each observation is correlated to an underlying latent variable, or ``state'',
and then derive the parameter transition operator between these states.

Some approaches directly view each observation as a state, and then focus on transitions between the observations in the time domain. For instance,
the work in~\cite{schodl2000video} treats this transition as an associated probability problem,
and other methods  construct a spatio-temporal autoregressive model (STAR) or position affine operator for this transition \cite{szummer1996temporal,kwatra2003graphcut}.

Differently, feature-based models capture the intrinsic law and underlying structures of the data by projecting the original data onto a low-dimensional feature space via feature extracted techniques, such as principle component analysis (PCA). 
G.~Doretto et al. \cite{saisan2001dynamic, doretto2003dynamic} model the evolution of the dynamic textured scenes as a linear dynamical system (LDS) under a Gaussian noise assumption. As a popular method in dynamic textures, LDS and its derivative algorithms have been successfully used for various dynamic texture applications \cite{doretto2003dynamic,saisan2001dynamic}. 
However, constraints are imposed on the types of motion and noise that can be modeled in LDS. For instance, it is sensitive to input variations due to various noise. Especially, it is vulnerable to non-Gaussian noise, such as missing data or occlusion of the dynamic scenes. Moreover, stability is also a challenging problem for LDS \cite{boots2007constraint}.

To tackle these challenges, the approach taken here is to explore an alternative method to model the DTs by appealing to the principle of sparsity. Instead of using the Principle Components (PCs) as the transition ``states'' in LDS, sparse coefficients over a learned dictionary are imposed as the underlying ``states''. In this way, the dynamical process of DTs exhibits a transition course of corresponding sparse events.
 These sparse events can be obtained via a recent technique on linear decomposition of data, called dictionary learning \cite{elad2006image,hawe2013separable}. Formally, these sparse representations $x\in \mathbb{R}^k$ to a signal $y\in \mathbb{R}^m$, can be written as
 $$y = Dx$$
 where $D\in \mathbb{R}^{m\times k}$ is a dictionary, and $x$ is sparse, i.e. most of its entries are zero or small in magnitude. That is, the signal $y$ can be sparsely represented only using a few elements from some dictionary $D$.

In this work, we start with a brief review of the dynamic texture model from the viewpoint of convex $\ell_2$ optimization, and then deduce a combined regression associated with several regularizations for a joint process---``states extraction'' and ``states transition''. Then we treat the solution of the above combined regression as an adaptive dictionary learning problem, which can achieve two distinct yet tightly coupled tasks--- efficiently reducing the dimensionality via sparse representation and robustly modeling the dynamical process. Finally, we cast this dictionary learning problem as the optimization of a smooth non-convex objective function, which is efficiently resolved via a gradient descent method.

\section{Adaptive Video Dictionary Learning}
\label{sec:03}
In this section, we start with a brief introduction to the linear dynamical systems (LDS)
model and develop an adaptive dictionary learning framework for sparse coding.

\subsection{Linear Dynamical Systems}
\label{sec:02}
%
Let us denote a given sequence of $(n+1)$ frames by $Y := [y_{0}, \ldots, y_{n}]
\in \mathbb{R}^{m\times (n+1)}$, where the time is indexed by $i = 0,1, \ldots, n$.
The evolution of a LDS is often described by the following two equations
\begin{equation}
\label{eq:lds}
	\left\{\!\!
	\begin{array}{rl}
	x_{i+1} &\!\!\!\! = A x_{i} + w_{i} \\
	y_{i} &\!\!\!\! = D x_{i} + v_{i},
	\end{array}
	\right.
\end{equation}
where $y_{i} \in \mathbb{R}^{m}$,
$x_{i} \in \mathbb{R}^{k}$, $w_{i} \in \mathbb{R}^{k}$ and $v_{i} \in \mathbb{R}^{m}$
denote the observation, its hidden state or feature, state noise, and observation noise, respectively.
The system is described by the dynamics
matrix $A \in \mathbb{R}^{k\times k}$, and the modeling matrix $D \in
\mathbb{R}^{m \times k}$.
Here we are interested in estimating the system parameters $A$ and $D$, together with the hidden states,
given the sequence of observations $Y$.

The problem of learning the
LDS \eqref{eq:lds} can be considered as a coupled linear regression problem \cite{boots2007constraint}.
Let us denote $X = [x_{0}, \ldots, x_{n}] \in \mathbb{R}^{k \times (n+1)}$,
$X_{0} = [x_{0}, \ldots, x_{n-1}] \in \mathbb{R}^{k \times n}$,
and $X_{1} = [x_{1}, \ldots, x_{n}] \in \mathbb{R}^{k \times n}$.
%
The system dynamics and
modeling matrix are expected to be caught by solving the
following
minimization problem,
\begin{equation}\label{FeatureBased1}
    \min_{A,D,X} \big\| X_{1} - A X_{0} \big\|_{F}^2 \quad s.t. \;
    \big\| Y - D X \big\|_{F}^2\leq \varepsilon,
\end{equation}
where $\varepsilon$ is a small positive constant.
In our approach, we assume that all observations $y_{i}$ admit a sparse representation
with respect to an unknown dictionary $D \in \mathbb{R}^{m \times k}$, i.e.
\begin{equation}
	y_{i} = D x_{i}, \qquad\text{for all}~i=0,1,\ldots,n,
\end{equation}
where $x_{i} \in \mathbb{R}^{k}$ is sparse. Without loss of generality, we
further assume that all columns of the dictionary $D$ have unit norm.
We then define the set
\begin{equation}
	\mathcal{S}(m,k) := \{ D \in \mathbb{R}^{m \times k} | \operatorname{ddiag}(D^{\top} D) = I_{k} \},
\end{equation}
where $\operatorname{ddiag}(Z)$ is the diagonal matrix whose entries on the diagonal
are those of $Z$, $I_{k}$ denotes the identity matrix.
The set $\mathcal{S}(m,k)$ is the product of $k$ unit spheres,
and is hence a $k(m-1)$ dimensional
smooth manifold.
Finally, by adopting the common sparse coding framework to problem \eqref{FeatureBased1},
we have the following minimization problem
\begin{equation}
\label{FeatureBased2}
    \min_{A,D,X} \big\| X_{1} - A X_{0} \big\|_{F}^2 + \mu_{1}
    \big\| Y - D X \big\|_{F}^2 + \mu_{2} \|X\|_{1},
\end{equation}
where $D\in \mathcal{S}(m,k)$, $\|\cdot\|_{F}$ denotes the Frobenius norm of matrices, and $\|\cdot\|_{1}$ is
the $\ell_{1}$ norm, which measures the overall sparsity of a matrix.
The parameter $\mu_{2} > 0$ weighs the sparsity measurement against the
residual errors.

\subsection{A Dictionary Learning Model for Dynamical Scene}
Solving the minimization problem as stated in Eq.~\eqref{FeatureBased2}
is a very challenging task.
%
In this work, we employ an idea similar to \emph{subspace identification methods} \cite{boots2007constraint}, which treat the state as a function of $(A,D)$.
Here, we confine ourselves to the sparse solution of an elastic-net problem, which is proposed in \cite{zou2005regularization}, as
\begin{equation}
\label{eq:elastic-net}
	x^{*} := \operatorname*{argmin}_{x \in \mathbb{R}^{k}}
	\tfrac{1}{2} \| y - D x \|_{2}^{2} + \lambda_{1} \|x\|_{1}
	+ \tfrac{\lambda_{2}}{2} \|x\|_{2}^{2},
\end{equation}
where $\lambda_1 > 0$ and $\lambda_2 > 0$ are regularization parameters, which play an
important role in ensuring stability and uniqueness of the solutions.
%
Let us define the set of indices of the non-zero entries of the solution $x^{*} = [
x_{1}^{*}, \ldots, x_{k}^{*}]^{\top} \in \mathbb{R}^{k}$ as
\begin{equation}
	\Lambda := \{i\in\{1,\ldots,k\} | x^{*}_{i} \neq 0\}.
\end{equation}
Then the solution $x^{*}$ has a closed-form expression as
\begin{equation}
\label{eq:enet}
	x_{y}^{*}(D) := \left( D_{\Lambda}^{\top} D_{\Lambda} - \lambda_{2}
	I_{m} \right)^{-1} \left( D_{\Lambda}^{\top} y - \lambda_{1}
	s_{\Lambda} \right),
\end{equation}
where $s_{\Lambda} \in \{ \pm 1\}^{|\Lambda|}$ carries the signs of
$x_{\Lambda}^{*}$, $D_{\Lambda}$ is the subset of $D$ in which
the index of atoms (rows) fall into support $\Lambda$.
Furthermore, it is known that the solution
$x_{y}^{*}(D)$ as given in \eqref{eq:enet} is a locally twice
differentiable function at $D$.
By an abuse of notation, we define
\begin{equation}
\begin{split}
	X_{0} \colon \mathcal{S}(m,k) \to &\, \mathbb{R}^{k \times n} \\
	D \mapsto &\, [x_{y_{0}\!}^{*}(D), \ldots, x_{y_{n-1}\!}^{*}(D)].
\end{split}
\end{equation}

In a similar way, $X_{1} \colon \mathcal{S}(m,k) \to \mathbb{R}^{k \times n}$ is defined.
Thus, the cost function reads as
%
\begin{equation}
\label{main_AVDL_unstable}
\begin{split}
	f \colon \mathbb{R}^{k \times k} \times \mathcal{S}(m,k) \to &\, \mathbb{R} \\
	(A, D) \mapsto &\, \tfrac{1}{2} \left\| X_{1}(D) - A X_{0}(D) \right\|_{F}^{2}.
\end{split}
\end{equation}

It is known that an LDS with the dynamic matrix $A$ is said to be stable, if the largest eigenvalue of $A$ is bounded by $1$ \cite{boots2007constraint}.
Let $\sigma$ be the largest eigenvalue of $A$, then $|\sigma|\leq \|A\|_F.$
Thus, we enforce the small $\sigma$ via imposing a penalty $\|A\|_F^2$ on \eqref{main_AVDL_unstable},
and then end up with the cost function as
\begin{equation}
\label{main_AVDL}
\begin{split}
	\widetilde{f} \colon \mathbb{R}^{k \times k} \times \mathcal{S}(m,k) \to &\, \mathbb{R} \\
	(A, D) \mapsto &\, f(A,D) + \tfrac{\gamma}{2}\|A\|_F^2,
\end{split}
\end{equation}

\subsection{Development of the Algorithm}
In this section, we firstly derive a gradient descent algorithm to minimize \eqref{main_AVDL}
and then discuss some details of the choice of the parameters in the final implementation.

%
We start with the computation of the first derivative of the sparse solution of
the elastic-net problem $x_{y}^{*}(D)$ as given in \eqref{eq:enet}.
Given the tangent space of $\mathcal{S}(m,k)$ at $D$ as
\begin{equation}
	T_{D}\mathcal{S}(m,k) := \{ X \in \mathbb{R}^{m \times k} | \operatorname{ddiag}(X^{\top} D) = 0 \},
\end{equation}
the orthogonal projection of a matrix $H \in \mathbb{R}^{m \times k}$ onto the
tangent space $T_{D}\mathcal{S}(p,n)$ with respect to the inner product
$\langle X,Y \rangle = \operatorname{tr}(X^{\top}Y)$ is given by
\begin{equation}
	\Pi_{D}(H) := H - D \operatorname{ddiag}(D^{\top} H).
\end{equation}
%
%
Let us denote $K := D_{\Lambda}^{\top} D_{\Lambda} - \lambda_{2} I_{k}$.
The first derivative of $x_{y}^{*}$ in the direction
$H \in T_{D}\mathcal{S}(m,k)$ is
\begin{equation}
\begin{split}
	\operatorname{D}x_{y}^{*}(D)H = &\,
	K^{-1} H_{\Lambda}^{\top} y - K^{-1}\!(D_{\Lambda}^{\top}
	H_{\Lambda} \\
    & + H_{\Lambda}^{\top} D_{\Lambda} )
	\cdot K^{-1} \left( D_{\Lambda}^{\top} y - \lambda_{1}
	s_{\Lambda} \right).
\end{split}
\label{eq:diriv_direction2}
\end{equation}

By the product structure of $\mathbb{R}^{k \times k} \times \mathcal{S}(m,k)$,
the Riemannian gradient of the function $\widetilde{f}$ is
\begin{equation}
\label{eq:main_diriv}
	\operatorname{grad} \widetilde{f}(A,D) = \left(\nabla_{\widetilde{f}}(A),
	\Pi_{D}\big(\nabla_{\widetilde{f}}(D)\big) \right).
\end{equation}
Here, the Euclidean gradient $\nabla_{\widetilde{f}}(A)$ of $\widetilde{f}$ with
respect to $A$ is computed as
\begin{equation}
\label{eq:diffA}
	\nabla_{\widetilde{f}}(A) = \left(A X_{0}(D) - X_{1}(D) \right)X_{0}(D)
	+ \gamma A,
\end{equation}
with $e_{i}$ being the $i$-th standard basis vector of $\mathbb{R}^{n}$.
Using the shorthand notation,
$r_{i} := D_{\Lambda_{i}}^{\top} y_{i} - \lambda_{1}s_{\Lambda_{i}}$,
$\Delta x_{i} := x_{y_{i}}^{*}(D)-A_{\Lambda_i}x_{y_{i-1}}^{*}(D)$, and $q_i := r_i\Delta x_{i}^{\top}$,
the Euclidean gradient $\nabla_{\widetilde{f}}(D)$ of $
\widetilde{f}$ with respect to $D$ is
%
\begin{equation}
\begin{split}
\label{eq:diff_D}
	& \nabla_{\widetilde{f}}(D) = \sum\limits_{i=1}^{n} y_{i} (\Delta x_{i})^{\top}
	K_{i}^{-1} - D_{\Lambda_{i}} K_{i}^{-1} (q_i+ q_i^{\top}) \\
    &\cdot K_{i}^{-1}- y_{i-1} (\Delta x_{i})^{\top} \!A_{\Lambda_i} \!(K_{i-1})^{-1} + D_{\Lambda_{i-1}} \\
	& \cdot (K_{i-1})^{-1} (A_{\Lambda_{i-1}}q_{i-1} + q_{i-1}^{\top}A_{\Lambda_{i-1}}^{\top}) (K_{i-1})^{-1}.
\end{split}
\end{equation}

For a gradient search iteration on manifolds,
we employ the following smooth curve on $\mathcal{S}(m,k)$ through $D \in \mathcal{S}(m,k)$
in direction $H \in T_{D}\mathcal{S}(m,k)$
\begin{equation}
\begin{split}
	\tau \colon &(-\lambda,\lambda) \to \mathcal{S}(m,k) \\
	t \mapsto & (D + t H) \big(\operatorname{ddiag}((D + t H)^{\top}
	(D + t H))\big)^{-\tfrac{1}{2}}
\end{split}
\end{equation}
with $\lambda >0$. It essentially normalizes all columns of $D + t H$. For a detailed overview on optimization on matrix manifold, refer to \cite{absil2009optimization}.

\begin{algorithm}[ht]
   \caption{Adaptive Video Dictionary Learning} \label{alg:AVDL}
\begin{algorithmic}[1]
   \STATE Training data $Y$
   \STATE Initialize the parameters $\lambda_1$,$\lambda_2$,$\gamma$, initial dictionary $D$, and initial transition matrix $A$.
  \FOR{$i=1,2,\ldots,T$}
   \STATE \emph{Sparse Coding Stage}\\
        $\quad$ Use Lasso algorithm to compute $x$ via\\
        $\quad$ $x\leftarrow\underset{x}{\min}\frac{1}{2}\|y-Dx\|_2^2 + \lambda_1\|x\|_1 + \frac{\lambda_2}{2} \|x\|_2^2$\\
        $\quad$ Compute the active set $\Lambda$ for each $x$. \\
   \STATE Compute the gradient of $\widetilde{f}(A,D)$ according to \eqref{eq:diffA} and \eqref{eq:diff_D}.
\vspace{6pt}
   \STATE Update the parameters $A$ and $D$
\begin{equation*}
\begin{split}
\label{eq:diriv_direction}
&A_i\leftarrow A_{i-1}-\rho_i \nabla_{\widetilde{f}}(A_{i-1}), \\
&D_i\leftarrow D_{i-1}-\rho_i \nabla_{\widetilde{f}}(D_{i-1}).
\end{split}
\end{equation*}
\ENDFOR
\RETURN {$A$ and $D$}
\end{algorithmic}
\end{algorithm}

Until now, we have computed the gradient of $\widetilde{f}$ as defined in \eqref{main_AVDL} with respect to its two arguments $D$ and $A$. An iterative scheme (such as the gradient descent method or conjugate gradient method) can be used to find the optimal $D$ and $A$, using the gradient expression above.
The procedure displayed in Algorithm \eqref{alg:AVDL} is the version of AVDL based on gradient descent procedure. The learning rate $\rho_i$ can be computed via the well-known backtracking line search method, similar to \cite{hawe2013separable}. Here, considering the high coherence among the temporal frames, we prefer non-redundant dictionary, that is, $k\ll m$ for the dictionary $D\in \mathbb{R}^{m\times k}$. For parameters $(\lambda_1,\lambda_2)$ in the elastic net, we put an emphasis on sparse solutions and choose $\lambda_2 \in (0,\frac{\lambda_1}{10})$, as proposed in \cite{zou2005regularization}.
\begin{table*}[htbp!]
\centering
\caption{\label{candle_sys} \small Synthesizing results on sequence of burning candle.
}
\begin{tabular}{c|c|c|c|c|c|c|c|c}
\hline
\multirow{2}{*}{Instance} & \multicolumn{3}{|c|}{LDS, (PCs)} & \multicolumn{5}{|c}{AVDL, $\gamma = 0.5$, (loops) } \\
\cline{2-9}
&64 & 128& 256& 1 & 50 & 100 & 200 & 400\\
\hline
Compression rate (\%) & 6.25              & 12.50             &  25.00            &     1.02          &3.29&   3.41 &  3.50    & 3.55  \\
\hline
$\sigma$         &  0.9802           &  0.9833           &    0.9849         &  1.78             & 1.06    & 0.9992  &  0.9994 & 0.9994 \\
\hline
$e_y$              & $1.35\times 10^5$ & $1.35\times 10^5$ & $1.35\times 10^5$ & $1.36\times 10^3$ &60.29 &$58.82$ & $55.97$ & 71.27  \\
\hline
$e_x$              &   $101.58$         &    $135.88$        &    $168.95$        & $3.75\times 10^4$ & 171.99& 75.52   & 61.96   & 46.18  \\
%

\hline
\end{tabular}
\end{table*}

\section{Numerical Experiments}
\label{sec:04}

\begin{figure}[htb]
  \centering
  \subfigure[Corrupted original sequence]{
    \label{fig1.1:subfig:a} 
    \includegraphics[width=1.58in]{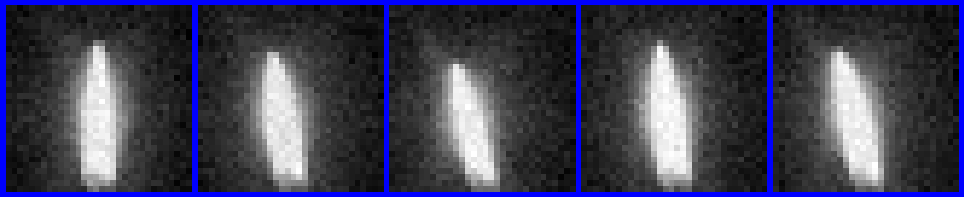}}
  \subfigure[Reconstructed sequence]{
    \label{fig1.1:subfig:b} 
    \includegraphics[width=1.58in]{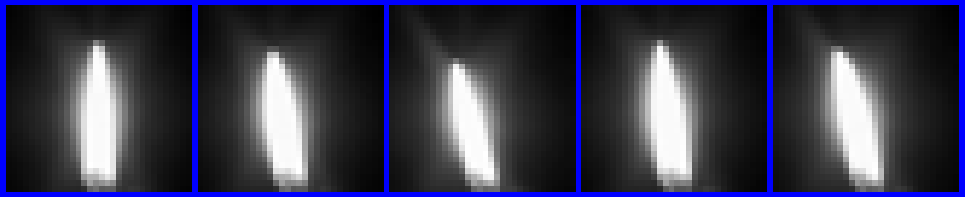}}
  \subfigure[Synthesized video using LDS and AVDL on DTs with Gaussian noise]{
    \label{fig1.1:subfig:c} 
    \includegraphics[scale=0.32]{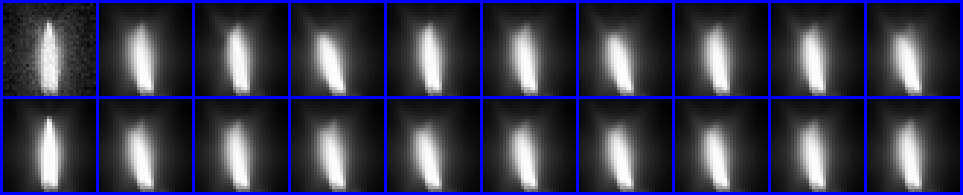}}
  \subfigure[Synthesized video using LDS and AVDL on DTs with missing data]{
    \label{fig1.1:subfig:b} 
    \includegraphics[width=3.14in]{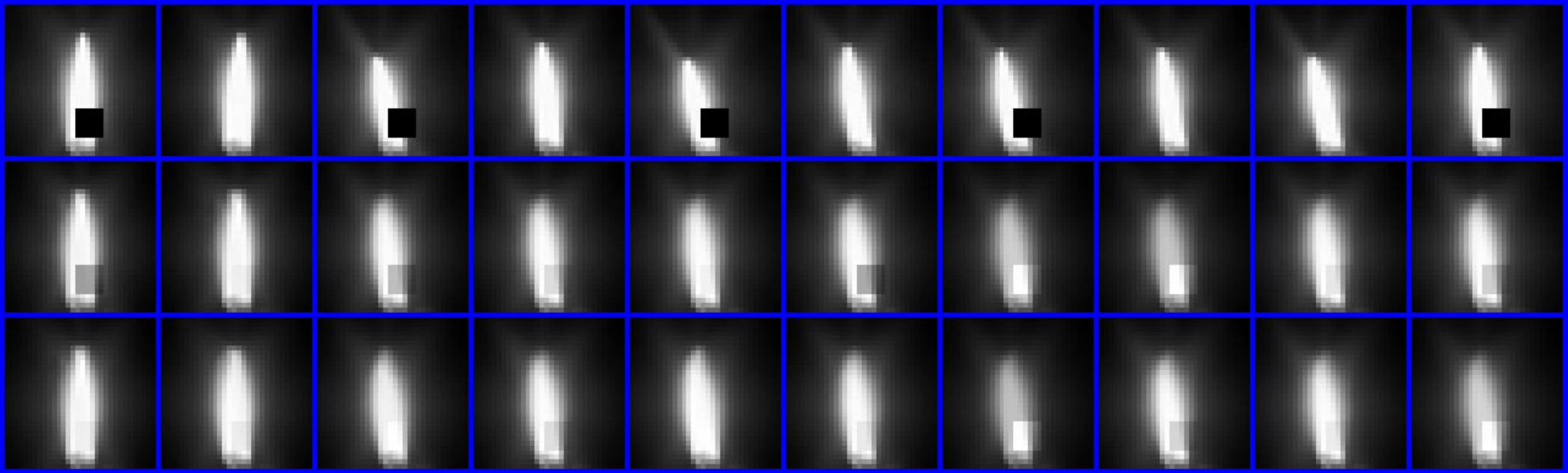}}
  \caption{ \small
  Reconstruction and synthesizing on the candle scene.
  (a), (b) are $(i = 1,64,128,512,1024){th}$ frame of the corrupted data by Gaussian noisy and the reconstructed data using AVDL, respectively. (c) The top row is the synthesized sequence using LDS (128PCs), and the bottom row is the synthesized sequence using AVDL ($(i = 2,1024,3072, 5120, \dots,20480){th}$ frame).
  (d) The top row is the sequence with missing data. The middle row the synthesized sequence using LDS, and the bottom row is the synthesized sequence using AVDL.
 }
 \label{fig_candle_sys_rec} 
\end{figure}

We carry out a few experiments on natural image sequences data, and demonstrate the practicality of the proposed algorithm. Our test dataset comprises of videos from DynTex++ \cite{ghanem2010maximum}, and data from internet sources (for instance, YouTube). Firstly, we show the performance on reconstruction and synthesizing with a grayscale video of burning candle, which is corrupted by Gaussian noise or occlusion. This video has 1024 frames with size of $32\times 32$, see figure~\ref{fig_candle_sys_rec}. The initial dictionary is $1024\times 512$. After the acquisition of the dictionary $D$ and the transition $A$, the synthesized data can be generated easily by $x_{i+1} = Ax_ix_i^{\top}x_i(x_i^\top x_i)^{-1}$, or more precisely, using a convex formulation
 $$\min_{x_{i+1}}\frac{1}{2}\|x_{i+1}-Ax_i\|_2^2+\lambda\|x_{i+1}\|_1.$$

Table~\ref{candle_sys} shows the performance of synthesizing on burning candle with Gaussian noise. The error pairs $(e_x, e_y)$  are defined as $e_{y}=\sum_{i}\|y_i-Dx_i\|$, $e_{x}=\sum_i\|x_{i+1}-Ax_i\|$, and the largest eigenvalue of $A$ is denoted by $\sigma$. The compression rate for AVDL is sparsity of $x$ to $m\times(n+1)$, and for LDS is number of PCs to $m$. Table~\ref{candle_sys} shows AVDL can obtain the stable dynamic matrix $A$ $(\sigma\leq1)$, smaller compression rate and smaller error $(e_x, e_y)$ of cost function \eqref{FeatureBased2}, by increasing the numbers of main loops in Algorithm 1.

Figure~\ref{fig_candle_sys_rec} $(a\sim c)$ is the visual comparison between LDS and AVDL. AVDL performs well on denoising against corruption by Gaussian noise. In the case of occlusion in figure~\ref{fig_candle_sys_rec} (d), random 50 frames of the 1024 burning candle video are corrupted by a $(6\times7)$ rectangle. The length of both synthesizing data is 1024, based on first frame of the burning candle. $87.01\%$ of the synthesizing data from LDS are corrupted by this rectangle, but $9.47\%$ for AVDL.
\begin{table}[htb!]
\begin{flushleft}
\centering
\caption{\small DT recognition rates for videos with occlusion.
}
\label{table:recognition}
\begin{tabular}
{lllll}
\hline \small
Occlusion rate (\%) &\small 0 &\small 5 &\small 15 &\small 30\\
\hline\small
\small LDS-NN (128PCs)  &\small 69.72 &\small 45.00 &\small 25.14 &\small 14.17 \\

\hline\small
\small AVDL-SRC          &\small 70.28 &\small 64.72 &\small 44.44 &\small 22.36\\

\end{tabular}
\end{flushleft}
\end{table}
The second experiment is about scenes classification on DynTex++, which contains DTs from 36 classes. Each class has 100 subsequences of length 50 frames with $50\times 50$ pixels.
20 videos are randomly chosen in each class and total 720 videos are used for our experiments. 
Classification for LDS is performed using the Martin distance with a nearest-neighbor classifier on its parameters pair $(A,D)$ \cite{saisan2001dynamic}. Another classifier is AVDL associated with the sparse representation-based classifier (SRC)  \cite{wright2009robust,ghanem2010sparse},
in which the class of a test sequence is determined by the smallest reconstruction error $e_{y}$
and transition error $e_{x}$.
Table~\ref{table:recognition} provides the recognition results with increasing occlusion rates for test data. Compared to LDS with nearest-neighbor classifier (LDS-NN), Table~\ref{table:recognition} shows the proposed AVDL with SRC (AVDL-SRC) performs better while the test videos are corrupted by increasing occlusion.

\section{Conclusions}
\label{sec:05}
This paper
proposes an alternative method, called AVDL, to model the dynamic process of DTs. In AVDL, the sparse events over a dictionary are imposed as transition states. The proposed method show a robust performance for synthesizing, reconstruction and recognition on DTs corrupted by Gaussian noise. Especially, AVDL exhibits more powerful in the case of test data with non-Gaussian noise, such as occlusion. One possible future extension is to learn a dictionary for large scale DT sequences based on AVDL.


\bibliographystyle{IEEEbib}

\end{document}